\UseRawInputEncoding
\documentclass[lettersize,journal]{IEEEtran}
\pdfoutput=1
\usepackage{amsmath, amsfonts, amssymb, amscd, mathtools, physics}
\usepackage{algorithmic}
\usepackage{array}
\usepackage{textcomp}
\usepackage{stfloats}
\usepackage{url}
\usepackage{verbatim}
\usepackage{graphicx}
\def\BibTeX{{\rm B\kern-.05em{\sc i\kern-.025em b}\kern-.08em
    T\kern-.1667em\lower.7ex\hbox{E}\kern-.125emX}}
\usepackage{balance}

\usepackage[utf8]{inputenc} 
\usepackage[T1]{fontenc}    
\usepackage{caption} 
\usepackage{hyperref}
\hypersetup{hidelinks}
\usepackage{url}            
\usepackage{booktabs}       
\usepackage{nicefrac}       
\usepackage{microtype}      
\usepackage{xcolor}         

\usepackage{dsfont}
\usepackage{orcidlink}
\usepackage{color}
\usepackage{cleveref}
\usepackage{ulem}

\usepackage{xcolor}
\definecolor{light-gray}{gray}{0.95}
\newcommand{\code}[1]{\colorbox{light-gray}{\texttt{#1}}}

\begin{document}

\title{\textit{lucie}: An Improved Python Package for Loading Datasets from the UCI Machine Learning Repository}
\date{\today}

\author{
    Kenneth Ge\orcidlink{0009-0000-5044-4433}, %
    Phuc Nguyen\orcidlink{0000-0001-9993-8434}, %
    and Ramy Arnaout\orcidlink{0000-0001-6955-9310}%
\thanks{
This work was supported by the Gordon and Betty Moore Foundation and by the NIH under grants R01HL150394, R01HL150394-SI, R01AI148747, and R01AI148747-SI.}

\thanks{Kenneth Ge is a research intern in the Department of Pathology at Beth Israel Deaconess Medical Center (BIDMC), and is a student at Carnegie Mellon University. Phuc Nguyen is with the Department of Pathology at BIDMC, Boston, MA 02215. Ramy Arnaout (email: rarnaout@bidmc.harvard.edu) is with the Department of Pathology and the Division of Clinical Informatics, Department of Medicine, BIDMC and with Harvard Medical School, Boston, MA 02215.}
}

\maketitle
\begin{abstract}
The University of California--Irvine (UCI) Machine Learning (ML) Repository (UCIMLR) is consistently cited as one of the most popular dataset repositories, hosting hundreds of high-impact datasets. However, a significant portion, including 28.4\% of the top 250, cannot be imported via the \textit{ucimlrepo} package that is provided and recommended by the UCIMLR website. Instead, they are hosted as .zip files, containing nonstandard formats that are difficult to import without additional ad hoc processing. To address this issue, here we present \textit{lucie}---\uline{l}oad \uline{U}niversity \uline{C}alifornia \uline{I}rvine \uline{e}xamples---a utility that automatically determines the data format and imports many of these previously non-importable datasets, while preserving as much of a tabular data structure as possible. \textit{lucie} was designed using the top 100 most popular datasets and benchmarked on the next 130, where it resulted in a success rate of 95.4\% vs. 73.1\% for \textit{ucimlrepo}. \textit{lucie} is available as a Python package on PyPI with 98\% code coverage.
\end{abstract}


\tableofcontents

\section{Introduction}\label{sec:intro}

\begin{figure}
    \centering
    \includegraphics[width=0.9\columnwidth]{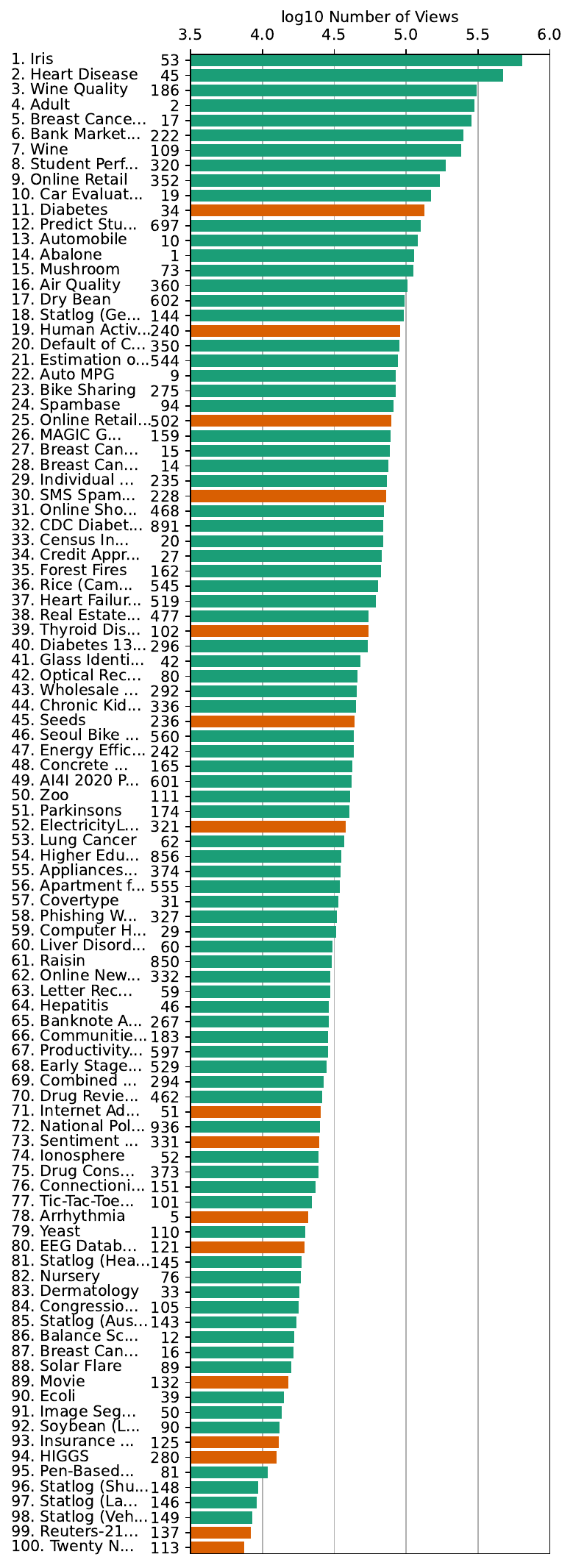}
    \caption{The 100 most popular UCIMLR datasets, ordered by number of views and presented using a log10 scale. Green=datasets that \textit{ucimlrepo} successfully imports; red=\textit{ucimlrepo} failures. The UCIMLR ID\# is displayed along with the name and rank. Note the correlation between success and popularity (more red bars toward the bottom).}
    \label{fig:datasets}
\end{figure}

Dataset accessibility is critical to advancement of machine learning research, across disciplines \cite{51055, dey_proceedings_2023, miao_minimum_2024, MAYFIELD201717, chinn_enriching_2023, 8317828}. The UCI ML Repository~\cite{uci2024} (UCIMLR) is considered one of the most popular machine-learning dataset repositories~\cite{badr2019}. However, many of the datasets it contains are old, containing custom formats that cannot be imported via the built-in \textit{ucimlrepo} Python package or API~\cite{ucimlrepo2024} or via the ``Import to Python'' button on the UCIMLR (Fig.~\ref{fig:datasets}). Although undoubtedly useful, UCI datasets are known to often contain missing fields, missing headers, not-a-number (NaN) columns, and most importantly, nonstandard data formats. Among the top 100 most popular datasets, 16 are not importable using \textit{ucimlrepo}. Among less popular datasets, the proportion that are not importable is even higher, possibly because datasets that are easier to use get more downloads. For datasets ranked 101-250, over a third cannot not be imported via \textit{ucimlrepo} (see Results).

A small fraction of these ``dirty''~\cite{reuters} datasets have been cleaned and mirrored on other dataset repositories such as Kaggle, Huggingface, or various GitHub repositories. Examples include Diabetes (UCMLR ID \#34) on Kaggle~\cite{kaggle_diabetes}, the Reuters-21578 Text Categorization Collection (\#347) on HuggingFace~\cite{huggingface_reuters}, and Movie (\#132) on GitHub~\cite{github_movies}. However, mirroring still leaves three major problems:

\begin{enumerate}
    \item The vast majority of the non-importable datasets still lack cleaned mirror versions,
    \item Importing these datasets via mirrors still requires manual identification and an ad-hoc per-dataset approach, and
    \item The work required to track down datasets via mirrors or other sites diminishes the ``all-in-one-place'' value of the UCIMLR.
\end{enumerate}
 
In our own work, we noticed that many of the problematic datasets, although differing from each other in data format and layout, share many common patterns in how they were constructed. We exploited these patterns to create \textit{lucie}---\uline{l}oad \uline{UCI} \uline{e}xamples---a smart-import tool that follows a nearly identical interface as \textit{ucimlrepo} and automatically guesses the data format to import previously non-importable datasets. Then, if possible, it tries to coerce the data into a dataframe using the popular Python \textit{pandas} package, just as \textit{ucimlrepo} does.

\section{Methods}

\subsection{Code and Datasets}

\textit{ucimlrepo} version 0.0.7 was installed using pip (\code{pip install ucimlrepo}). We determined dataset popularity rankings based on the results given by the ``sort by num views, desc'' criteria on the UCIMLR (accessed August 26, 2024). After saving these results, we attempted to import each of the top 100 using \textit{ucimlrepo}. This identified 16 that were non-importable: UCMLR IDs \#5, 34, 51, 102, 113, 121, 125, 132, 137, 228, 236, 240, 280, 321, 331, and 502 (alphabetically).

We downloaded these datasets using the download-as-zip function on the UCIMLR, and manually identified the following patterns in their construction:
\begin{itemize}
\item 5 datasets contained tabular data, in the form of ``.data'' files, ``.xlsx'' files, or some other clearly tabular data format;
\item 4 datasets contained many ``.txt'' files, some of which contained metadata and some of which contained tabular data;
\item 4 datasets contained highly irregular data formats, such as HTML or XML with hyperlinks representing a relational database;
\item 2 datasets contained nested archives, where each folder represented a column and each file was an entry; and
\item 1 dataset contained only extension-free files, some of which contained metadata and some of which contained tabular data stored in a delimiter-separated format.
\end{itemize}

\noindent We concluded that if these patterns occurred at least once, there was a good chance they would occur again, i.e. in datasets beyond the top 100.

We constructed our import algorithm to be as general as possible, while minimizing any false positives (which we define to be interpreting extraneous files as data). \textit{lucie}'s import function returns a dictionary, with keys representing the names of any files identified as tabular data (including archives), and values being corresponding \textit{pandas} dataframes. If that is not possible, as a fallback, it instead returns a dictionary representing the directory structure and its contained files for any identified data archives; however, in practice, no such failure mode was detected among the top 100 datasets. Finally, we also modified the \code{fetch\_ucirepo} function to return a tuple, with the first element being ``uci'' or ``custom'', representing whether the dataset was imported using the built-in function vs. our custom import algorithm, respectively. 

\begin{figure}
    \centering
    \includegraphics[width=1\linewidth]{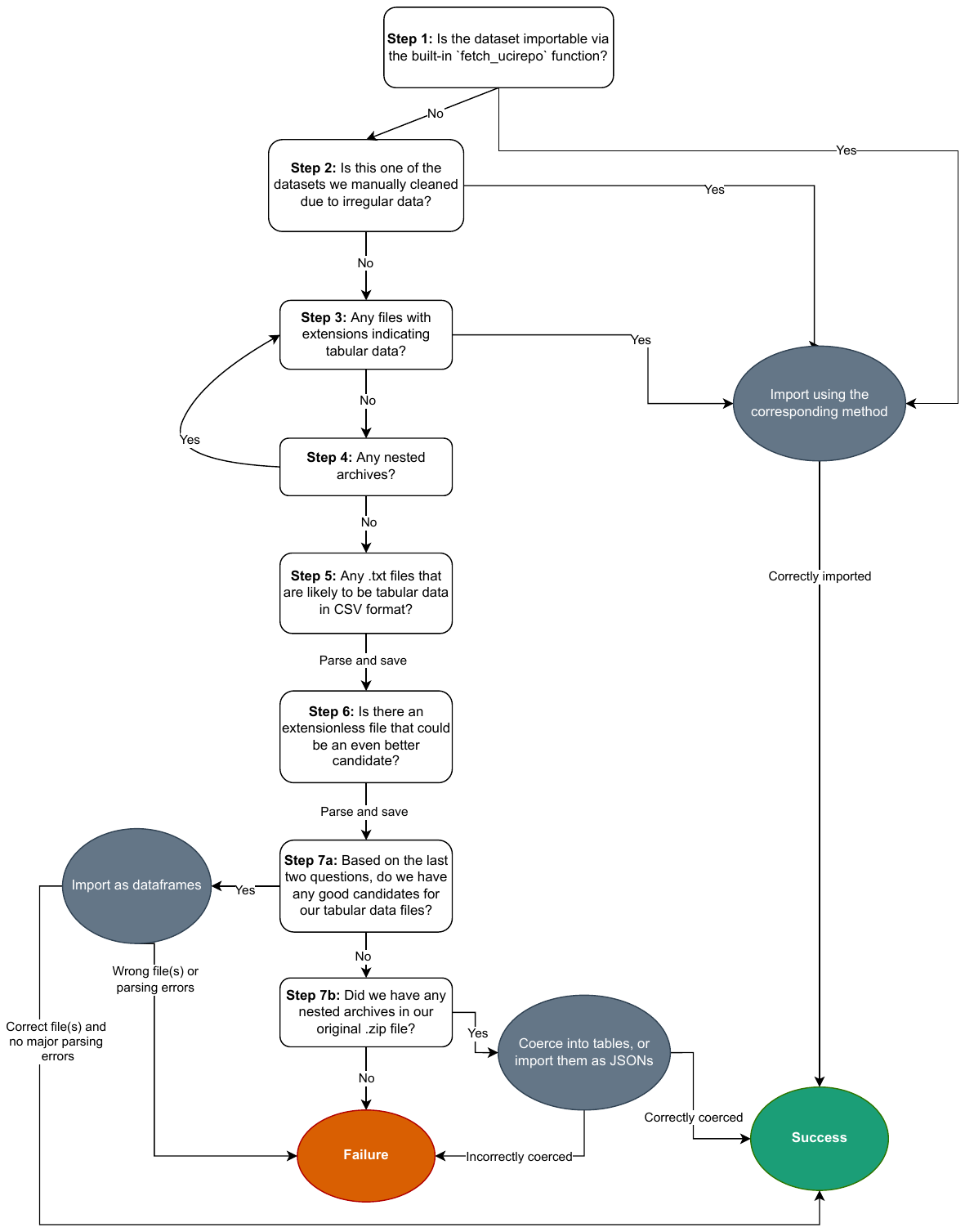}
    \caption{Flowchart of the \textit{lucie} algorithm, with question boxes corresponding to the steps described in the text. The green oval indicates success, the red oval indicates failure, and the gray ovals indicate an action.}
    \label{fig:algorithm}
\end{figure}

\subsection{Algorithm (Fig.~\ref{fig:algorithm})}

\textbf{Step 1}
The import algorithm first queries the UCIML API to determine if a given dataset can already be imported using the built-in \code{fetch\_ucirepo} function. If so, it returns a tuple containing the string ``uci'' and the result. If not, we use BeautifulSoup~\cite{beautifulsoup} to scrape the UCIMLR for the download link for the corresponding ``.zip'' file. Then, we extract this archive. 

\textbf{Step 2}
For four repositories containing highly irregular data formats (EEG with ID \#121, Diabetes \#34, Movie \#132, and Reuters-21578 \#137), we found pre-cleaned mirrors and execute custom import code; \textit{lucie} uses these, treating them as special cases. 

\textbf{Step 3}
For the rest of the datasets, we start by looking for any files with extensions that clearly identify them as tabular data, such as ``.data'', ``.xlsx'', ``.tsv'', ``.csv'', etc. If found, we read them all and return them in the dictionary format described above. Since some of these files may have an irregular shape (i.e. some rows have more columns than others), we use a custom \code{read\_csv} function that removes the requirement that the data be rectangular.

\textbf{Step 4}
If no such files are found, we then look for any nested archives. Some datasets contain multiple nested archives; for example, the ``Twenty Newsgroups'' dataset contains a ``mini'' version that is a sample from the larger dataset. So, to determine which archive to try first, we sort them by name, using their edit distance from the word ``data.'' These nested archives then get extracted and processed recursively by going back to Step 3. If any tabular data is found, we exit and return it.  

\textbf{Step 5}
If Step 2 is unsuccessful, we look through all ``.txt'' files and try to determine if they are likely to be tabular data disguised as plain text. We do this by trying various common delimiters (comma, semicolon, and tab) and determining how much of the resulting table is NaN. If the data is ``perfect''---i.e. it contains multiple columns, no NaN data, and a regular shape---then we return the resulting dataframe. Otherwise, we keep trying other ``.txt'' files to find the ones with the lowest proportion of NaN entries and the most regular shape. In the event of a tie, we save all of them.

\textbf{Step 6}
We repeat the same process as in Step 5 for files without an extension, but due to the increased likelihood of false positives, we only accept a data file if it has a lower proportion of NaN entries compared to all the other files we have tried so far. Furthermore, we only save the single best file, rather than multiple files like we did in Step 5. 

\textbf{Step 7}
If no likely candidates are identified, we go back to any nested archive. Then, we try to coerce it into tables, by treating nested folders as columns and nested files as entries. If this is unsuccessful, we read the archive as a JSON dictionary mirroring the directory structure (with keys representing file and values representing content), and return this instead. 

The flow chart in Fig.~\ref{fig:algorithm} provides a more comprehensive view of the logical structure.

\section{Availability}

\textit{lucie} is available via PyPI (\code{pip install lucie}) and on GitHub at https://github.com/ArnaoutLab/lucie. Forks and contributions are welcome. 

\section{Results}

The \textit{lucie} algorithm was designed using manually identified patterns found in the top 100 most-popular datasets and tested on datasets ranked 101-250. Dataset imports were considered successful if the following criteria held:
\begin{itemize}
\item The algorithm returned a meaningful result, and did not crash, error out, or return ``None;''
\item The first few rows of the data files seemed to contain meaningful data upon visual inspection; and
\item The resulting dataframes had many columns and rows (as opposed to just one or a few rows or being empty).
\end{itemize}

\noindent If all three criteria were not met, the import was considered a failure. 

\begin{figure} 
    \centering
    \includegraphics[width=0.95\columnwidth]{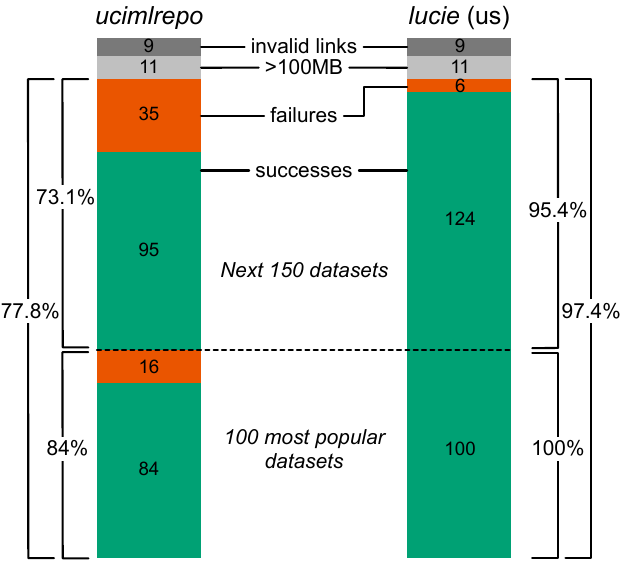}
    \caption{Percent success for \textit{ucimlrepo} (left) and \textit{lucie} (right). Green=successful imports; red=failed imports. Percents are shown for the 100 most popular datasets, 130 of the 150 next most popular datasets (separated by dotted line), and these 100+130=230 datasets; the remaining 20 were excluded for having broken/external links (dark gray) and large datasets/timeouts (light gray). Integers in the bars are no. datasets. The 100 most popular datasets, on which \textit{lucie} was developed, are on the bottom.}
    \label{fig:results}
\end{figure}

As a baseline, \textit{ucimlrepo} succeeded on 84 of the top 100 and failed on the remaining 16 (Fig.~\ref{fig:results}). In contrast, \textit{lucie} succeeded on all 100 (which effectively served as its training set). As a test, we asked how many of the datasets ranked 101-250 in popularity could be imported by \textit{ucimlrepo} vs. \textit{lucie}. We found that 20 of the datasets in this test set failed for reasons unrelated to importing: 9 did not have valid download links (UCMLR IDs \#683, 752, 770, 892, and 465), caused download errors internal to UCMLR (\#516), or had a size of 0 bytes (\#611, 613, and 346) and 11 were over 100MB which we excluded due to size-related timeouts (UCMLR IDs \#442, 920, 344, 791, 779, 773, 256, 231, 226, 203, and 164). We therefore excluded these 9+11=20 datasets from the test set and asked how many of the remaining 150-20=130 datasets could be imported by each utility. \textit{ucimlrepo} successfully imported 95 of these 130 (73.1\%), while \textit{lucie} successfully imported 124 out of these 130 (95.4\%) (Fig.~\ref{fig:results}). \textit{lucie} failed on UCMLR IDs \#7, 25, 130, 180, 432, and 470.

We also wrote \textit{pytest} tests scripts, which are included in both the GitHub repository and in the PyPI package. We tested all of the common dataset types, and for rare dataset types that we did not encounter in practice, we created synthetic test datasets to demonstrate that \textit{lucie} would support these functions if they were to come up. Our tests result in a code-coverage rate of 98\%; the coverage report is included in the GitHub repository. Please see the repository for more information on the tests.

\section{Discussion}

The breathtaking progress of AI/ML and related fields depends on the ready availability of widely used/high-quality datasets for benchmarking and investigation \cite{offord2023, yoong_benefits_2022, ferreira_label-free_2022}. The UCIMLR plays a major role. Frictionless availability of data is critical to open and collaborative science more generally, especially valuable for difficult problems and in times of urgency or crisis~\cite{openscience2023, lakhani2013, bicer_koios_2011, cooperation2021, omicron2022, 52002}. However, these datasets are only useful if they can be easily imported. The diversity and distributed nature of datasets and their contributors understandably and inevitably lead to failure in the default import process, which new code development, such as \textit{lucie}, can ameliorate. We have demonstrated that \textit{lucie} successfully imports the 100 most-popular datasets (as of late 2024), rescuing some failures from the default import method (\textit{ucimlrepo}), and has a high success rate with non-importable datasets outside of this range, as evidenced by 95.4\% success rate on datasets 101-250 (excepting broken/outside links and datasets over 100MB, which time out). Future work can help increase the success rate by incorporating more data formats and manually adding cleaned datasets wherever necessary. For example, one potential low hanging fruit could be a more sophisticated way of dealing with ``.Z'' archives, which caused issues in some of the six datasets in our test set that failed to import.

Future work could analyze the test set more closely, since the analysis of the success rate presented here was based only on the three criteria mentioned and not on, for example, a comprehensive, row-by-row check. An improvement would be to study each dataset more in-depth, and determine manually if any files were missed, or conversely, if \textit{lucie} imported any false positives. Finally, any such codebase requires maintenance, especially as additional datasets are added to the UCIMLR and increase in popularity, and benefits from extension, to make ever more datasets easily useable by the research community. To this end, \textit{lucie} is open source and distributed under the MIT license; any future contributions are welcome. Issues can be left at the GitHub repository.

\section{Conflict of Interest}

The authors declare that they have no conflict of interest.

\hfill

\bibliographystyle{IEEEtran}
\bibliography{ref}


\vskip -2\baselineskip plus -1fil

\begin{IEEEbiographynophoto}{Kenneth Ge} is a research intern at the Beth Israel Deaconess Medical Center in Boston, MA, USA. He is also an undergraduate studying Computer Science at Carnegie Mellon University in Pittsburgh, PA, USA. He has previous research experience and publications in Human Computer Interaction and Accessible Computing, as well as Machine Learning. 
\end{IEEEbiographynophoto}%

\vskip -2\baselineskip plus -1fil

\begin{IEEEbiographynophoto}{Phuc Nguyen} received a B.S. degree in Physics from the University of Texas at Austin in 2010, and a Ph.D. degree in Physics from the University of Texas at Austin in 2017. From 2017 to 2023, he worked as a postdoctoral researcher in physics at the University of Maryland in College Park, the City University of New York, the University of Haifa in Israel and Brandeis University, with research interest spanning from high-energy physics (especially the AdS/CFT correspondence) to general relativity to quantum information science. He is a postdoctoral researcher at Beth Israel Deaconess Medical Center in Boston, MA, USA, with research interest focussing on information theory and machine learning.

\end{IEEEbiographynophoto}%

\vskip -2\baselineskip plus -1fil

\begin{IEEEbiographynophoto}{Ramy Arnaout} received an S.B. degree in biology from the Massachusetts Institute of Technology in 1997, a D.Phil. (Ph.D.) degree in Mathematical Biology (Zoology) from the University of Oxford in 1999, and an M.D. degree with honors from Harvard Medical School in 2003. He is Associate Professor of Pathology at Harvard Medical School and Associate Director of the Clinical Microbiology Laboratories at the Beth Israel Deaconess Medical Center, both in Boston, MA, USA. His past and present research interests include entropy, diversity, immunomics, metagenomics, computational pathology, informatics, artificial intelligence, and machine learning. Dr. Arnaout is a member of the College of American Pathologists.\end{IEEEbiographynophoto}

\end{document}